\documentclass[lettersize,journal]{IEEEtran}
\usepackage{amsmath,amsfonts}
\usepackage{algorithmic}
\usepackage{algorithm}
\usepackage{array}
\usepackage[caption=false,font=normalsize,labelfont=sf,textfont=sf]{subfig}
\usepackage{textcomp}
\usepackage{stfloats}
\usepackage{url}
\usepackage{verbatim}
\usepackage{graphicx}
\usepackage{cite}
\usepackage{multirow}
\usepackage{booktabs}
\usepackage[table,xcdraw]{xcolor}

\begin{document}

\title{Forgetting to Remember: A Scalable Incremental Learning Framework for Cross-Task Blind Image Quality Assessment}

\author{Rui Ma,~Qingbo Wu,~\IEEEmembership{Member,~IEEE,}~King Ngi Ngan,~\IEEEmembership{Life Fellow,~IEEE,}~Hongliang Li,~\IEEEmembership{Senior Member,~IEEE,}~Fanman Meng,~\IEEEmembership{Member,~IEEE,}~Linfeng Xu,~\IEEEmembership{Member,~IEEE}
\thanks{This work was supported in part by the National Natural Science Foundation of China under Grant 61971095, Grant 61831005, and Grant 61871087. (Corresponding authors: Qingbo Wu; Fanman Meng.) The authors are with the School of Information and Communication Engineering, University of Electronic Science and Technology of China, Chengdu 611731, China (e-mail: ruima@std.uestc.edu.cn; qbwu@uestc.edu.cn; knngan@cuhk.edu.hk; hlli@uestc.edu.cn; fmmeng@uestc.edu.cn; lfxu@uestc.edu.cn).}}

\markboth{Journal of \LaTeX\ Class Files,~Vol.~14, No.~8, August~2021}%
{Shell \MakeLowercase{\textit{et al.}}: A Sample Article Using IEEEtran.cls for IEEE Journals}

\IEEEpubid{0000--0000/00\$00.00~\copyright~2021 IEEE}

\maketitle

\begin{abstract}
Recent years have witnessed the great success of blind image quality assessment (BIQA) in various task-specific scenarios, which present invariable distortion types and evaluation criteria. However, due to the rigid structure and learning framework, they cannot apply to the cross-task BIQA scenario, where the distortion types and evaluation criteria keep changing in practical applications. This paper proposes a scalable incremental learning framework (SILF) that could sequentially conduct BIQA across multiple evaluation tasks with limited memory capacity. More specifically, we develop a dynamic parameter isolation strategy to sequentially update the task-specific parameter subsets, which are non-overlapped with each other. Each parameter subset is temporarily settled to \textit{Remember} one evaluation preference toward its corresponding task, and the previously settled parameter subsets can be adaptively reused in the following BIQA to achieve better performance based on the task relevance. To suppress the unrestrained expansion of memory capacity in sequential tasks learning, we develop a scalable memory unit by gradually and selectively pruning unimportant neurons from previously settled parameter subsets, which enable us to \textit{Forget} part of previous experiences and free the limited memory capacity for adapting to the emerging new tasks. Extensive experiments on eleven IQA datasets demonstrate that our proposed method significantly outperforms the other state-of-the-art methods in cross-task BIQA. The source code of the proposed method is available at https://github.com/maruiperfect/SILF.
\end{abstract}


\section{Introduction}
\begin{figure*}[!t]
	\centering
	\includegraphics[width=7in]{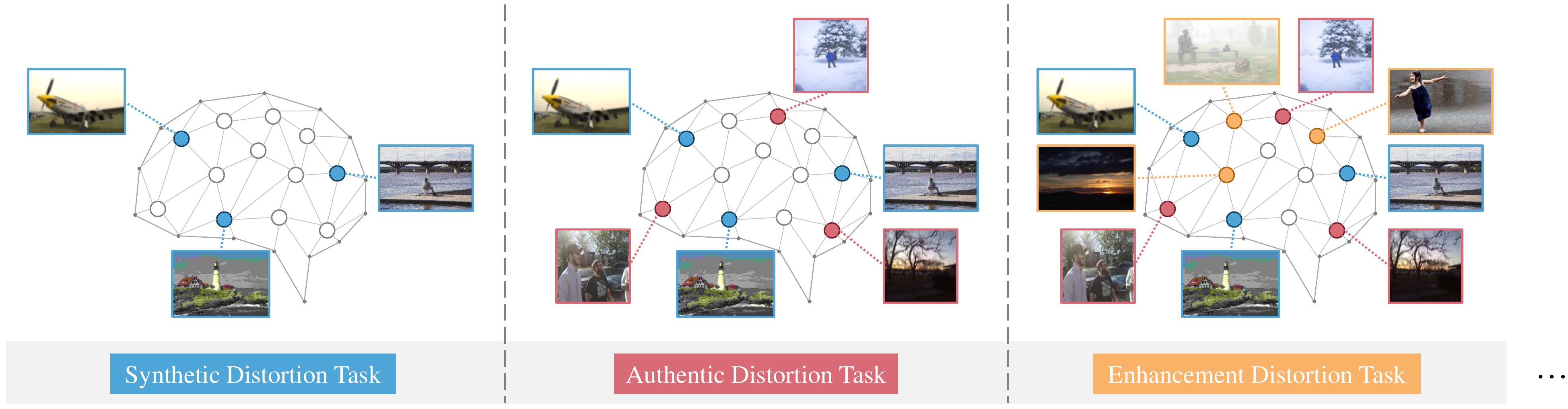}
	\caption{Cross-task blind image quality assessment process. The model sequentially evaluates different types of BIQA tasks. Zoom in to observe better the distorted details of the images for various tasks.}
	\label{figure_1}
\end{figure*}

\IEEEPARstart{B}{lind} image quality assessment (BIQA) aims to evaluate the perceptual quality of an image where the prior knowledge of its reference image and distortion type is unavailable. Due to the absence of pristine reference and the unpredictability of various distortions in the open environment, BIQA is highly desired in many practical applications, such as image quality monitoring, camera tuning, network adaptation, and so on. Recently, many BIQA methods \cite{zhang2018predicting, zhang2020learning, jiang2017optimizing, yang2019sgdnet, jiang2022single, jiang2022underwater, jiang2017blique, xu2022quality, zhai2020comparative, huang2021image} have achieved impressive performance in diverse task-specific evaluations, including synthetic, authentic, and enhancement distortion evaluation tasks. However, in practical applications, the distortion artifacts and evaluation criteria may keep changing with the task scenario, which brings new demand for cross-task BIQA. More specifically, we require a general-purpose model to sequentially learn the perceptual preference across different types of BIQA tasks, which mimics the continual knowledge accumulation of the human, as shown in Fig. \ref{figure_1}.

\IEEEpubidadjcol

The most straightforward approach is to train a separate model for every new task. Despite the best performance for the cross-task scenario, this method requires a lot of storage space and computing resources, which is challenging to apply in practice. Another solution is to use only one static model and fine-tune the old model to adapt to new tasks, whose storage and computation overhead would not grow with the increasing new tasks. Unfortunately, the model tends to forget old knowledge after learning a new task. This phenomenon, known as catastrophic forgetting \cite{mccloskey1989catastrophic}, where the model tends to change the weights necessary for previous tasks to adapt to new tasks, occurs when the model learns multiple tasks sequentially. 

Recently, several pioneering works \cite{zhang2021task, zhang2022continual, liu2022liqa} have been proposed to address the catastrophic forgetting problem in cross-task BIQA. Zhang et al. \cite{zhang2021task} propose a simple yet effective method for continually learning BIQA models by training task-specific batch normalization parameters for each task while keeping all pre-trained convolution filters fixed. Based on a shared backbone network, Zhang et al. \cite{zhang2022continual} add a prediction head for a new dataset and enforce a regularizer to allow all prediction heads to evolve with new data while resisting old data catastrophic forgetting. Limited by the hard parameter sharing of the backbone, these methods \cite{zhang2021task, zhang2022continual} are hard to apply to the BIQA tasks with significant differences. LIQA \cite{liu2022liqa} employs a generator conditioned on the distortion type and the quality score to generate pseudo features, which serve as a memory replayer when learning new tasks. Despite the success of adapting to significantly different BIQA tasks, LIQA requires additional computational and memory resources for generating and reusing the paired pseudo features and labels. Unlike the aforementioned hard parameter sharing and replay strategies, in our previous work \cite{ma2021remember}, we proposed a remember and reuse network (R\&R-Net) to separate all model parameters into predefined subsets according to the specified task capacity. This rigid parameter isolation strategy efficiently avoids catastrophic forgetting with limited memory. However, in the meantime, it also prevents R\&R-Net from learning additional tasks which exceed the specified task capacity.

This paper extends R\&R-Net with the dynamic parameter isolation strategy and scalable memory unit.
Specifically, we set two pruning ratios for each preset task: the first and second pruning ratios. The first pruning aims to release the redundant parameters of the current task for subsequent preset tasks. The second pruning enables the scalability of our model to learn additional tasks that exceed the preset task capacity. After two prunings, we obtain two candidate models corresponding to the current task, which we call the maximum and minimum models. During fine-tuning, we train the two models alternately. In practical applications, if there are no additional tasks, the preset tasks can employ their corresponding maximum models to achieve the best performance. Otherwise, the preset tasks can exploit their related minimum models to free the memory of the scalable memory unit to learn the additional tasks. By means of negligible sacrifice on the preset tasks, our model can increase the task capacity without expanding the memory capacity.

Compared to our previous work, the additional contributions of this paper are summarized in the following:

\begin{enumerate}
	\item{We propose a scalable incremental learning framework (SILF). By gradually and selectively pruning unimportant neurons from previously settled parameter subsets, we develop a scalable memory unit, which enables us to forget part of previous experiences and free the limited memory capacity for adapting to emerging new tasks. In this way, we effectively extend the task capacity of the model without increasing its memory capacity.}
	
	\item{We conduct extensive experiments to evaluate the proposed method on eleven IQA databases, including the synthetic, authentic, and enhancement distortion evaluation tasks. The results demonstrate that our proposed method significantly outperforms many state-of-the-art methods, even for the additional tasks.}
\end{enumerate}

The rest of this paper is organized as follows. Section II reviews the related work. Section III presents the proposed method. Section IV introduces experimental results and their analysis. Finally, Section V concludes the paper.

\section{Related Work}
This section briefly reviews representative BIQA and incremental learning methods closely related to our work.

\subsection{Blind Image Quality Assessment}
Existing BIQA methods can be divided into opinion-aware BIQA and opinion-unaware BIQA, depending on whether opinion scores are required during the training process.

\textbf{Opinion-aware BIQA} methods learn regression models based on training images and subjective evaluations of the people involved to predict the perceptual quality of the test images. Traditional opinion-aware BIQA methods extract hand-crafted quality-aware features and then convert these features into quality scores employing a regression function. Natural scene statistic (NSS) based methods \cite{gu2014using, li2016blind, gu2016blind, zhan2017structural, wu2017blind, moorthy2011blind, saad2012blind, mittal2012no} are the most common of these methods. Some learning-based methods \cite{ye2012unsupervised, zhang2015som}, which learn features from training data, have also been developed to overcome the limitations of hand-crafted features. The convolutional neural network (CNN) has recently achieved great success in various computer vision tasks, such as image classification, object detection, and semantic segmentation. Due to its robust feature representation capabilities, the CNN-based BIQA methods \cite{kang2014convolutional, kim2016fully, bosse2017deep, bare2017accurate, guan2017visual, zhang2018blind, yang2019cnn, zhu2020metaiqa, su2020blindly} have shown superior prediction performance over traditional methods. Inspired by the deep residual model \cite{he2016deep}, Bare et al. \cite{bare2017accurate} add two sum layers to a simple CNN network for BIQA. Bosse et al. \cite{bosse2017deep} modify VGGNet \cite{simonyan2014very} to learn local weights to measure the importance of the local quality of each image patch and employ weighted average patch aggregation as a pooling method. Zhang et al. \cite{zhang2018blind} propose a deep bilinear CNN-based (DB-CNN) BIQA model by conceptually modeling both synthetic and authentic distortions as two-factor variations by bilinear pooling. Inspired by deep meta-learning \cite{finn2017model}, Zhu et al. \cite{zhu2020metaiqa} propose an optimization-based meta-learning method for BIQA, which applies several distortion-specific NR-IQA tasks to learn prior knowledge of multiple distortions in images.

\textbf{Opinion-unaware BIQA} methods do not require subjective human opinions for training.
Nature image quality evaluator (NIQE) \cite{mittal2012making} extracts local features from an image and fits the feature vectors to a multivariate Gaussian model (MVG). It then predicts the quality of the test image based on the distance between its MVG model and the MVG model learned from a corpus of new naturalistic images. Zhang et al. \cite{zhang2015feature} then extend NIQE to integrated local NIQE (IL-NIQE). The IL-NIQE model extracts five types of NSS features from a collection of original natural images and then applies them to learn an MVG model of original images as a reference model for predicting the quality of the image patches. Given a test image, the model evaluates the scores of its patches separately and then employs the mean of these scores as the score of the test image. Wu et al. \cite{wu2015highly} propose a local pattern statistics index (LPSI) characterized by its good generality to various distortions. Liu et al.\cite{liu2020blind} propose natural scene statistics and perceptual characteristics-based quality index (NPQI) by investigating the human brain's NSS and perceptual features for visual perception.

\subsection{Incremental Learning}
Incremental learning has been a long-standing research problem in machine learning, focusing on learning efficient models from sequentially arriving data. Existing incremental learning methods can be summarily grouped into the following categories: namely, regularization-based, replay-based, dynamic architecture-based, and parameter isolation-based.

\textbf{Regularization-based} methods introduce regularization terms into the loss function to penalize network parameter changes when learning the current task to prevent catastrophic forgetting. Learning without forgetting (LwF) \cite{li2017learning} attempts to prevent model parameters from changing while training the current task by applying a cross-entropy loss regularized by a distillation loss \cite{hinton2015distilling}. Elastic weight consolidation (EWC) \cite{kirkpatrick2017overcoming} limits the change of essential parameters for previous tasks by imposing a quadratic penalty term that encourages weights to move in directions with low Fisher information. Schwarz et al. \cite{schwarz2018progress} then propose online EWC (OEWC), which optimizes the EWC by reducing the cost of estimating the Fisher information matrix.

\textbf{Replay-based} methods store representative examples of previous tasks in a replay buffer \cite{rebuffi2017icarl, belouadah2019il2m, liu2020mnemonics} or train a generative model to generate samples of previous tasks \cite{shin2017continual, wu2018memory, xiang2019incremental}. The model is jointly trained by combining the stored data with the data from the current task. Rebuffi et al. \cite{rebuffi2017icarl} select the samples closest to the average sample of each task as the replay subset. Belouadah et al. \cite{belouadah2019il2m} introduce second memory to store statistics of previous tasks. Another popular strategy is to train a generative adversarial network (GAN) \cite{goodfellow2014generative}. GAN can be employed to synthesize exemplars for previous tasks. Shin et al. \cite{shin2017continual} propose a cooperative dual model architecture consisting of deep generative and task-solving models. Xiang et al. \cite{xiang2019incremental} propose an incremental learning strategy based on conditional adversarial networks.

\textbf{Dynamic architecture-based} methods adapt new incoming tasks by modifying/expanding the network structure \cite{rusu2016progressive, yoon2017lifelong, hung2019compacting, li2019learn, rajasegaran2019random, singh2020calibrating}. Progressive neural networks (PNN) \cite{rusu2016progressive} develop a new column of the entire neural network for each task and transfer previously learned task features via lateral connections. Dynamically expandable networks (DEN) \cite{yoon2017lifelong} expand the network capacity by the number of units required for each task arrival. Rajasegaran et al. \cite{rajasegaran2019random} propose a randomized path selection method called RPS-Net, which applies a randomized path selection method for each task.

\textbf{Parameter isolation-based} methods assign diverse parameters to different tasks to prevent catastrophic forgetting \cite{fernando2017pathnet, serra2018overcoming, mallya2018packnet, mallya2018piggyback, ebrahimi2019uncertainty}. Serra et al. \cite{serra2018overcoming} introduce a task-based intricate attention mechanism that preserves the information of previous tasks without affecting the learning of the current task. Mallya et al. \cite{mallya2018packnet} propose PackNet for learning multiple tasks using iterative pruning. Ebrahimi et al. \cite{ebrahimi2019uncertainty} propose an incremental learning formulation with Bayesian neural networks, which utilizes uncertainty predictions to conduct incremental learning. Essential parameters are either fully preserved by a stored binary mask or modified according to their uncertainty for learning new tasks.

\begin{algorithm}[h]
	\caption{Scalable Task Incremental Learning Algorithm}
	\label{algorithm_1}
	\begin{algorithmic}[1]
		\REQUIRE
		Datasets: $\left\{ \mathcal{D}_t \right\}_{t=1}^{n+k}$, 
		
		task capacity: $n$, scalable task capacity: $k$, 
		
		first pruning ratios: $\mathcal{P} = \left\lbrace p_{1}, p_{2}, \cdots, p_{n} \right\rbrace$,
		
		second pruning ratios: $\mathcal{P^{'}} = \left\lbrace p_{1}^{'}, p_{2}^{'}, \cdots, p_{n}^{'} \right\rbrace$,
		
		\ENSURE
		Initial model: $f_0$, initial mask: $\mathcal{M}_0$,
		
		\STATE \# Sequential tasks
		\FOR {task $T_t \in \left\{ T_1, T_2, \cdots, T_{n+k} \right\}$}
		
		\IF {$t=1$}
		\STATE \# First task
		\STATE Train $f_0$ on $\mathcal{D}_1$, get $f_1$;
		\STATE Sequentially prune $f_1$ with ratios $p_{1}$ and $p_{1}^{'}$, get masks $\mathcal{M}_1$ and $\mathcal{M}_1^{'}$;
		\STATE Apply $\mathcal{M}_1$, $\mathcal{M}_1^{'}$ to $f_1$, successively, and fine-tune several epochs on $\mathcal{D}_1$;
		\STATE Save $f_1$, $\mathcal{M}_1$, and $\mathcal{M}_1^{'}$;
		
		\ELSIF {$1 < t \le n$}
		\STATE \# Preset tasks
		\STATE Apply Algorithm \ref{algorithm_2} to exploit the relevance between $T_t$ and $\left\{ T_1, T_2, \cdots, T_{t-1} \right\}$, get $f_{t-1}^{'}$;
		\STATE Train $f_{t-1}^{'}$ on $\mathcal{D}_t$, get $f_t$;
		\STATE Sequentially prune $f_t$ with ratios $p_{t}$ and $p_{t}^{'}$, get masks $\mathcal{M}_t$ and $\mathcal{M}_t^{'}$;
		\STATE Apply $\mathcal{M}_t$, $\mathcal{M}_t^{'}$ to $f_t$, successively, and fine-tune several epochs on $\mathcal{D}_t$;
		\STATE Save $f_t$, $\mathcal{M}_t$, and $\mathcal{M}_t^{'}$;
		
		\ELSIF {$n < t \le n+k$}
		\STATE \# Additional tasks
		\STATE Apply Algorithm \ref{algorithm_2} to exploit the relevance between $T_t$ and $\left\{ T_1, T_2, \cdots, T_{t-1} \right\}$, get $f_{t-1}^{'}$;
		\IF {$t = n+1$}
		\STATE Initialize $\mathcal{M}_{t-1}^{'}$ to $\mathcal{M}_t$;
		\ELSE
		\STATE Initialize $\mathcal{M}_{t-1}$ to $\mathcal{M}_t$;
		\ENDIF
		\STATE Apply $\mathcal{M}_t$ to $f_{t-1}^{'}$, then fine-tune several epochs on $\mathcal{D}_t$, get $f_t$;
		\STATE Save $f_t$ and $\mathcal{M}_t$;
		\ENDIF
		
		\ENDFOR
		
	\end{algorithmic}
\end{algorithm}

\begin{figure*}[t]
	\centering
	\includegraphics[width=7in]{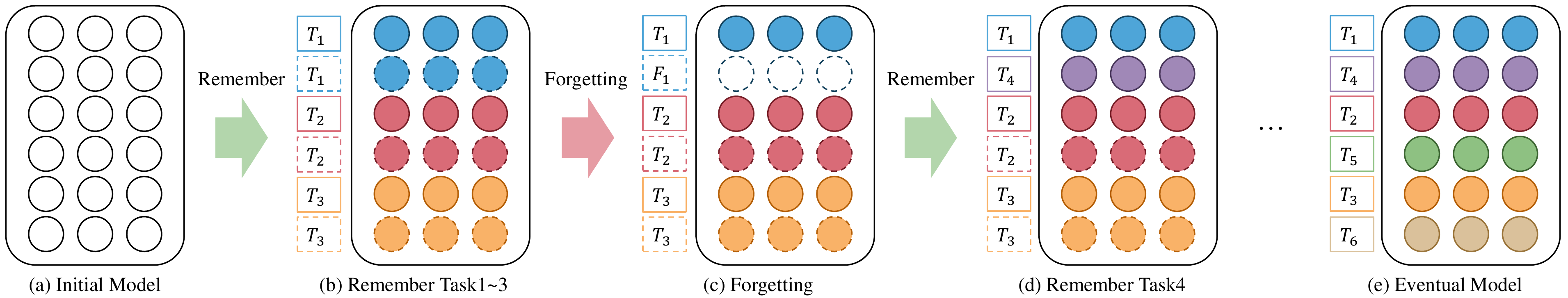}
	\caption{The training process of SILF, where (a)-(b) belong to the preset stage and (c)-(d) belong to the additional stage. The numerous circles in those rounded rectangles indicate the parameters of the model. The circle's color corresponds to the task index it belongs to (e.g., blue corresponds to task 1). The dashed circles indicate that these parameters are available for the additional stage.}
	\label{figure_2}
\end{figure*}

\begin{algorithm}[t]
	\caption{Task Relevance Guided Reuse Algorithm}
	\label{algorithm_2}
	\begin{algorithmic}[1]
		\REQUIRE
		Current dataset: $\mathcal{D}_t=\left\{ X_t, Y_t \right\}$, 
		
		previous model: $f_{t-1}$, masks: $\left\{ \mathcal{M}_1, \mathcal{M}_2, \cdots, \mathcal{M}_{t-1} \right\}$,
		
		\FOR {$i \in \left\{ 1, 2, \cdots, t-1 \right\}$}
		\STATE Apply $\mathcal{M}_i$ to $f_{t-1}$, get $f_i=f_{t-1} \cdot \mathcal{M}_i$;
		\STATE Input $X_t$ into $f_i$, get scores $S_{t, i}=f_i(X_t)$;
		\STATE Calculate $SRCC_{t, i}$;
		\STATE Calculate the reuse ratio $R_{t, i}$;
		\STATE Keep important parameters for $T_t$ in $T_i$ with a ratio of $R_{t, i}$, and prune the rest;
		\ENDFOR
	\end{algorithmic}
\end{algorithm}

\section{The Proposed Method}
A scalable incremental learning framework (SILF) for cross-task BIQA is proposed in this section. The training process of SILF can be divided into two stages, as shown in Fig. \ref{figure_2}. The first is the preset stage. The initial model progressively remembers preset tasks, as shown in Fig. \ref{figure_2}(a)-(b). With a sequentially remember and reuse strategy, our proposed network not only avoids the catastrophic forgetting problem in cross-task BIQA but also improves the performance of BIQA by selectively reusing some parameters of the learned tasks. The second is the additional stage, which is shown in Fig. \ref{figure_2}(c)-(e). Thanks to the scalable learning strategy, our model can adapt to additional tasks by forgetting some of the parameters of the learned tasks.

Without loss of generality, our work follows a sequential task incremental learning setting, a familiar setting in incremental learning. The training details of SILF are summarized in Algorithm \ref{algorithm_1}. The schematic diagram of a $4*4$ filter update process in SILF during the training process is shown in Fig. \ref{figure_3}. In the remainder of this section, we present our approach in the form of sequential tasks.

\subsection{Remember Preset Tasks}
\textbf{Input and Initialization}: Given initial model $f_0$ pre-trained on ImageNet \cite{deng2009imagenet}, initial mask $\mathcal{M}_0$ filled with $1$, and datasets $\mathcal{D}$. We pre-set the task capacity as $n$, the scalable task capacity as $k$, the first pruning ratios as $\mathcal{P}$, and the second pruning ratios as $\mathcal{P^{'}}$. We divide $\mathcal{D}$ into datasets of preset tasks $\left\{ \mathcal{D}_t \right\}_{t=1}^{n}$ and datasets of additional tasks $\left\{ \mathcal{D}_t \right\}_{t=n+1}^{n+k}$.

\textbf{Task 1}: First, we train $f_0$ on $\mathcal{D}_1$ and optimize all parameters of $f_0$. The optimized model is defined as $f_1$.

Second, we successively perform two sequential prunings on $f_1$. The pruning ratio of the two prunings is set to $p_1$ and $p_{1}^{'}$ respectively. We obtained masks $\mathcal{M}_1$ and $\mathcal{M}_1^{'}$ after two pruning sessions.

\begin{figure*}[t]
	\centering
	\includegraphics[width=7in]{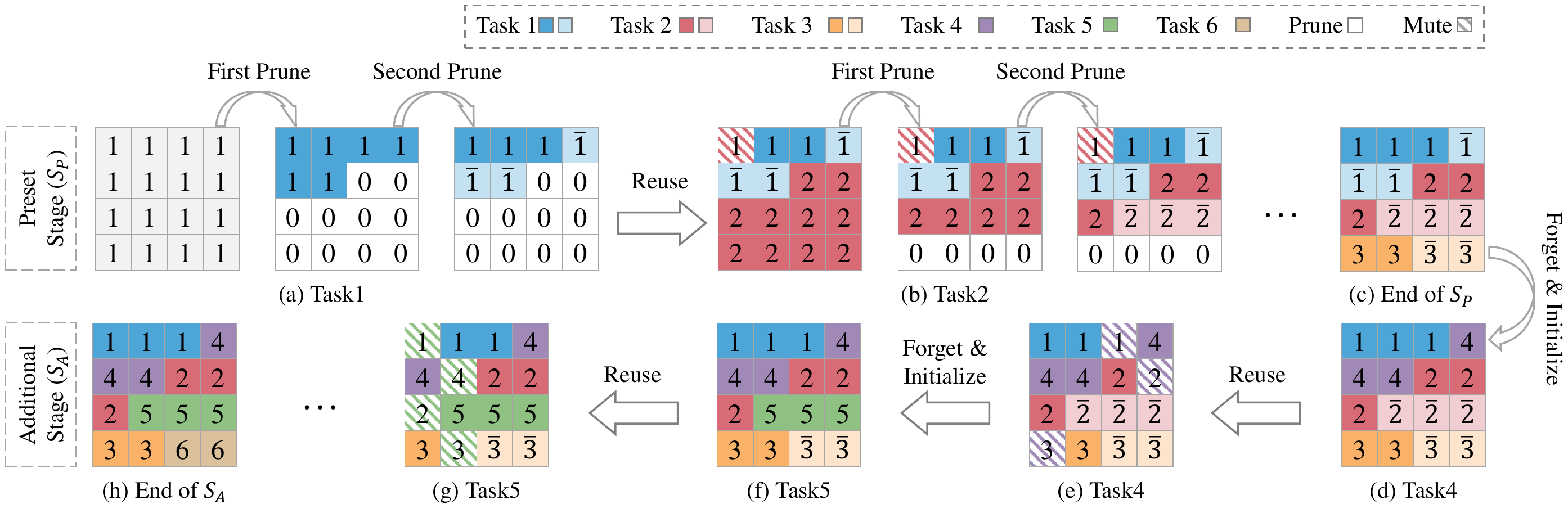}
	\caption{The evolution of a $4 \times 4$ filter during training. Each square in the filter corresponds to a number. The number equal to 0 indicates that the parameters of the position are pruned, and the number not equal to 0 represents the task label. The shaded square in the figure indicates that the parameters at this position are not included in the calculation of the current task (i.e., "Mute").}
	\label{figure_3}
\end{figure*}

Fig. \ref{figure_3}(a) shows the change process of a filter during the training of $T_1$. The left side of Fig. \ref{figure_3}(a) shows the initial mask filled by $1$, which belongs to $\mathcal{M}_0$. The middle of Fig. \ref{figure_3}(a) shows the result after the first pruning, which belongs to $\mathcal{M}_1$, with the pruned part filled by 0. The right side of Fig. \ref{figure_3}(a) is the result after the second pruning, which belongs to $\mathcal{M}_1^{'}$, part of the parameters belonging to $T_1$ is pruned, and the pruned part is changed from dark to light blue.

Take the first pruning process of a convolutional layer in $f_1$ as an example. We first sort all parameters of this layer by absolute value and then calculate the pruning threshold based on the pruning ratio $p_1$. Parameters above the threshold are considered essential for $T_1$, and the corresponding value in the mask is set to $1$ (i.e., the current task index). Parameters below the threshold are considered unimportant for $T_1$, and the corresponding value in the mask is set to $0$. Similar operations are performed at each layer of $f_1$ to obtain $\mathcal{M}_1$. The second pruning is done after the first, and the pruning process is similar to the first.

Third, we cyclically train the model after the first and second pruning. We first apply  $\mathcal{M}_1^{'}$ to $f_1$ and fine-tune several epochs on $\mathcal{D}_1$, then apply $\mathcal{M}_1$ to $f_1$ and fine-tune several epochs on $\mathcal{D}_1$. After several cycles of the above training steps, we obtain the scalable model $f_1$. Finally, we save the fine-tuned model $f_1$, masks $\mathcal{M}_1$ and $\mathcal{M}_1^{'}$.

\textbf{Task t ($1 < t \le n$)}: We call the learning stage of $T_t(1 < t \le n)$ the preset stage, denoted by $S_P$. Before the training of $T_t$, we apply Algorithm \ref{algorithm_2} to explore the relevance between $T_t$ and $\left\{ T_1, T_2, \cdots, T_{t-1} \right\}$. For $T_i \in \left\{ T_1, T_2, \cdots, T_{t-1} \right\}$, we first apply $\mathcal{M}_i$ to $f_{t-1}$:
\begin{equation}
	\label{equation_1}
	f_i=f_{t-1} \cdot M_i
\end{equation}
then we input the training set of $\mathcal{D}_t$ into $f_i$ to obtain the prediction scores:
\begin{equation}
	\label{equation_2}
	S_{t, i}=f_i(X_t)
\end{equation}
where $X_t$ is the training set of $\mathcal{D}_t$. Then $SRCC_{t, i}$ (Spearman Rank order Correlation Coefficient) is calculated based on predicted scores $S_{t, i}$ and ground truth scores $Y_t$. The reuse ratio between $T_t$ and $T_i$ is defined as:
\begin{equation}
	\label{equation_3}
	R_{t,i}=
	\begin{cases}
		1 + \lambda \cdot SRCC_{t,i},  & \text{if } SRCC_{t,i} < 0 \\
		1,  & \text{otherwise.}
	\end{cases}
\end{equation}
where $\lambda$ is a constant, which we set to 0.5 in this paper. Equation (\ref{equation_3}) aims to quantify the parameter reuse ratio of one task by considering its relevance to another task. More specifically, we tend to increase the reuse ratio of the current task from relevant knowledge and decrease the reuse ratio of the current task from irrelevant knowledge. Since previous tasks' data are unavailable, we use the previous model's performance on current data, i.e., SRCC, to measure the task relevance in continual learning, where a better performance means higher task relevance and vice versa.

For each $T_i \in \left\{ T_1, T_2, \cdots, T_{t-1} \right\}$, $T_t$ reuses the parameters belonging to $T_i$ in proportion $R_{t, i}$, and the rest of the parameters belonging to $T_i$ are muted. In the subsequent training of $T_t$, we temporarily ignore the muted part. The remaining training steps for $T_t$ are similar to $T_1$ and will not be repeated here.

Fig. \ref{figure_3}(b) shows the update process of the filter during the training of $T_2$. Before the training of $T_2$, we first apply Algorithm \ref{algorithm_2} to calculate the reuse ratio $R_{2,1}$ between $T_2$ and $T_1$. Then we mute some parameters belonging to $T_1$ according to $R_{2,1}$ (shaded squares in Fig. \ref{figure_3}(b) are muted). The next steps are similar to Task 1 and will not be repeated here.

\subsection{Remember Additional Tasks by Forgetting}
\textbf{Task t ($n < t \le n+k$)}: We call the learning stage of $T_t(n < t \le n+k)$ the additional stage, denoted by $S_A$. In preset stage $S_P$, the second pruned mask of $T_t(1 < t \le n)$ is saved as $\mathcal{M}_{t}^{'}$. The locations of $mask=t$ and $mask=\bar{t}$ in $\mathcal{M}_{t}^{'}$ indicate that the parameters at these locations belong to $T_t(1 < t \le n)$, where the locations of $mask=\bar{t}$ indicate that the parameters corresponding to these locations can be used for subsequent additional tasks. In Fig. \ref{figure_3}, dark and light squares of the same color belong to the same task, with light colors indicating availability for subsequent additional tasks. Take Fig. \ref{figure_3}(a) as an example, the dark blue squares correspond to $mask=1$, and the light blue squares correspond to $mask=\bar{1}$.

Before the training of $T_t(n < t \le n+k)$, we first initialize $\mathcal{M}_t$ by forgetting. Specifically, we initialize $\mathcal{M}_{t-1}^{'}$ to $\mathcal{M}_t$ if $t=n+1$, and we initialize $\mathcal{M}_{t-1}$ to $\mathcal{M}_t$ if $t>n+1$. The initialization operation replaces the part of $\mathcal{M}_{t-1}$ or $\mathcal{M}_{t-1}^{'}$ where $mask=\overline{t-n}$ is with $mask=t$. Fig. \ref{figure_3}(c)-(d) show the initialization process of $M_4$ before the training of $T_4$. We initialize the part of $M_3^{'}$ with $mask=1$ to $mask=4$ so that we can utilize this part of the parameters to learn the additional $T_4$.

Then we apply Algorithm \ref{algorithm_2} to explore the relevance between $T_t(n < t \le n+k)$ and $\left\{ T_1, T_2, \cdots, T_{t-1} \right\}$. The operation of calculating the reuse ratio is the same as $T_t(1 < t \le n)$. It is not repeated here. Finally, we apply $\mathcal{M}_t$ to $f_t$ and fine-tune it on $\mathcal{D}_t$. Fig. \ref{figure_3}(h) shows the state of the mask at the end of $S_A$. After all the learning steps in this section, our proposed SILF completes the learning of $S_P$ and $S_A$.

\section{Experimental Results}
In this section, we first describe the experimental setups, including IQA datasets, performance evaluation metrics, and implementation details. Then we compare the performance of the proposed with the state-of-the-art methods. Finally, we conduct a series of ablation studies to compare the performance of different settings. 

\begin{table*}[]
	\begin{center}
		\caption{Summary of IQA datasets used in our experiments}
		\label{table_1}
		\setlength{\tabcolsep}{4.5mm}{
			\begin{tabular}{lccccccc}
				\toprule[1pt]
				\textbf{\multirow{2}{*}{Dataset}} & \textbf{Distort.} & \textbf{\multirow{2}{*}{Content}} & \textbf{\# Distort.} & \textbf{\# Distort.} & \textbf{Subjective} & \textbf{Subjective} & \textbf{\multirow{2}{*}{Year}} \\
				& \textbf{Type} &  & \textbf{Types} & \textbf{Images} & \textbf{Environment} & \textbf{Rating Type} & \textbf{}\\ \toprule[0.5pt]
				LIVE \cite{sheikh2006statistical} & \multirow{5}{*}{Synthetic} & 29 & 5 & 779 & Laboratory & DMOS & 2006 \\
				CSIQ \cite{larson2010most} & & 30 & 6 & 866 & Laboratory & DMOS & 2010 \\
				LIVE-MD \cite{jayaraman2012objective} & & 15 & 4 & 450 & Laboratory & DMOS & 2012 \\
				TID2013 \cite{ponomarenko2013color} & & 25 & 24 & 3,000 & Laboratory & MOS & 2013 \\
				KADID-10k \cite{lin2019kadid} & & 81 & 25 & 10,125 & Crowdsourcing & MOS & 2019 \\ \toprule[0.5pt]
				LIVE-CH \cite{ghadiyaram2015massive} & \multirow{3}{*}{Authentic} & 1,169 & N/A & 1,169 & Crowdsourcing & MOS & 2016 \\
				KONIQ-10k \cite{hosu2020koniq} & & 10,073 & N/A & 10,073 & Crowdsourcing & MOS & 2018 \\
				SPAQ \cite{fang2020perceptual} & & 11,125 & N/A & 11,125 & Laboratory & MOS & 2020 \\ \toprule[0.5pt]
				SHRQ-Aerial \cite{min2019quality}      & \multirow{4}{*}{Enhanced} & 30 & N/A & 240 & Laboratory & MOS & 2019 \\
				SHRQ-Regular \cite{min2019quality} & & 45 & N/A & 360 & Laboratory & MOS & 2019 \\
				IVIPC-DQA \cite{wu2019beyond} & & 206 & N/A & 1,236 & Laboratory & MOS & 2019 \\
				LIEQ \cite{zhai2021perceptual} & & 100 & N/A & 1,000 & Laboratory & MOS & 2021 \\ \toprule[1pt]
		\end{tabular}}
	\end{center}
\end{table*}

\subsection{Experimental Setups}
\subsubsection{Datasets}
The main experiments are conducted on the following datasets: 

\textbf{Synthetically distorted datasets,} including LIVE \cite{sheikh2006statistical}, CSIQ \cite{larson2010most}, LIVE-MD \cite{jayaraman2012objective}, TID2013 \cite{ponomarenko2013color}, and KADID-10k \cite{lin2019kadid}. LIVE-MD consists of two multi-distortion datasets with a total of 450 distorted images. KADID-10k is a recently published large-scale synthetic database containing 81 reference images degraded by 25 distortion types in 5 levels for 10,125 distorted images.

\textbf{Authentically distorted datasets,} including LIVE-CH \cite{ghadiyaram2015massive}, KONIQ-10k \cite{hosu2020koniq}, and SPAQ \cite{fang2020perceptual}. LIVE-CH contains 1162 images captured under highly diverse conditions by many camera devices. KONIQ-10k and SPAQ are two recently proposed large-scale datasets containing 10,073 and 11,125 authentic distortion images, respectively.

\textbf{Enhancement distorted datasets} including SHRQ \cite{min2019quality}, IVIPC-DQA \cite{wu2019beyond}, and LIEQ \cite{zhai2021perceptual}. SHRQ contains two subsets, regular and aerial image subsets, respectively. We treat them as two separate datasets, denoted by SHRQ-A and SHRQ-R, respectively. IVIPC-DQA contains 1,236 derained images generated from 6 single image rain removal algorithms. LIEQ includes 1,000 light-enhanced images generated from 100 low-light images using 10 low-light image enhancement algorithms. Please refer to Table \ref{table_1} for details of each dataset.

\subsubsection{Performance Evaluation Metrics}

We employ the commonly used metric SRCC \cite{wu2018perceptually} to evaluate BIQA performance. The larger the value of SRCC, the better the performance of BIQA.
Following the criteria of \cite{lopez2017gradient}, we evaluate the performance of cross-task BIQA from three perspectives: average accuracy, average forgetting, and average plasticity in the SRCC. Average accuracy is the average SRCC of all BIQA tasks at the end of their incremental learning process, which is defined as:
\begin{equation}
	\label{equation_6}
	A_T=\frac{1}{T}\sum_{i=1}^{T}SRCC_{T,i}
\end{equation}
Average forgetting measures how much information the model has forgotten about previous tasks, which is formulated as:
\begin{equation}
	\label{equation_7}
	F_T=\frac{1}{T-1} \sum_{i=1}^{T-1} max_{t \in \left \{ 1, ...,T \right \} } (SRCC_{t, i}-SRCC_{T, i})
\end{equation} 
Average plasticity aims to evaluate the ability of the model to adapt to new tasks, which is defined as:
\begin{equation}
	\label{equation_8}
	P_T=\frac{1}{T}\sum_{i=1}^{T}SRCC_{i,i}
\end{equation}
where $T$ is the total number of learned tasks, $SRCC_{m,n}$ is the SRCC of the $n_{th}$ task after learning the first $m$ tasks sequentially. 

\begin{table*}[]
	\begin{center}
		\caption{Cross-Task BIQA performance comparison (Average Accuracy) on task sequence I}
		\label{table_2}
		\setlength{\tabcolsep}{3mm}{
			\begin{tabular}{l|ccccccc}
				\toprule[1pt]
				& \multicolumn{7}{c}{\textbf{Tasks}}                                                                                                                                                                                            \\ \cline{2-8} 
				\multirow{-2}{*}{\textbf{Method}} & \textbf{$1^{st}$(LIVE-CH)}                    & \textbf{$2^{nd}$(CSIQ)}                    & \textbf{$3^{rd}$(LIEQ)}                    & \textbf{$4^{th}$(KONIQ)}                    & \textbf{$5^{th}$(LIVE-MD)}                    & \textbf{$6^{th}$(SHRQ-A)}                    & \textbf{Mean}                 \\ \hline
				SL                                & \textbf{0.8529}                        & \textbf{0.8953}                        & \textbf{0.8990}                        & \textbf{0.8990}                        & \textbf{0.9105}                        & \textbf{0.9191}                        & \textbf{0.8960}                        \\
				LWF                               & 0.8260                                 & 0.3796                                 & 0.3565                                 & 0.4025                                 & 0.3543                                 & 0.2989                                 & 0.4363                                 \\
				PNN                               & 0.6809                                 & 0.3159                                 & 0.2640                                 & 0.3764                                 & 0.2726                                 & 0.3044                                 & 0.3690                                 \\
				R\&R-Net (3 tasks)                & 0.8163                                 & 0.8665                                 & 0.8667                                 & /                                      & /                                      & /                                      & /                                      \\
				R\&R-Net (6 tasks)                & 0.7910                                 & 0.8359                                 & 0.8560                                 & {\color[HTML]{FE0000} \textbf{0.8594}} & 0.8206                                 & 0.8252                                 & 0.8314                                 \\
				NO-RL                             & 0.8039                                 & 0.4663                                 & 0.4455                                 & 0.4001                                 & 0.4008                                 & 0.2775                                 & 0.4657                                 \\
				NO-RR                             & {\color[HTML]{FE0000} \textbf{0.8494}} & {\color[HTML]{FE0000} \textbf{0.8786}} & {\color[HTML]{FE0000} \textbf{0.8694}} & 0.8389                                 & {\color[HTML]{3166FF} \textbf{0.8591}} & {\color[HTML]{3166FF} \textbf{0.8420}} & {\color[HTML]{3166FF} \textbf{0.8562}} \\
				Proposed (SILF)                   & {\color[HTML]{3166FF} \textbf{0.8472}} & {\color[HTML]{3166FF} \textbf{0.8779}} & {\color[HTML]{3166FF} \textbf{0.8680}} & {\color[HTML]{3166FF} \textbf{0.8569}} & {\color[HTML]{FE0000} \textbf{0.8765}} & {\color[HTML]{FE0000} \textbf{0.8645}} & {\color[HTML]{FE0000} \textbf{0.8652}} \\ \toprule[1pt]
			\end{tabular}
		}
	\end{center}
\end{table*}

\begin{table*}[]
	\begin{center}
		\caption{Cross-Task BIQA performance comparison (Average Accuracy) on task sequence II}
		\label{table_3}
		\setlength{\tabcolsep}{3mm}{
			\begin{tabular}{l|ccccccc}
				\toprule[1pt]
				& \multicolumn{7}{c}{\textbf{Tasks}}                                                                                                                                                                                                                                                           \\ \cline{2-8} 
				\multirow{-2}{*}{\textbf{Method}} & \textbf{$1^{st}$(LIVE)}                             & \textbf{$2^{nd}$(SPAQ)}                             & \textbf{$3^{rd}$(KADID)}                             & \textbf{$4^{th}$(IVIPC-DQA)}                             & \textbf{$5^{th}$(TID2013)}                             & \textbf{$6^{th}$(SHRQ-R)}                             & \textbf{Mean}                          \\ \hline
				SL                                & \textbf{0.9711}                        & \textbf{0.9420}                        & \textbf{0.9131}                        & \textbf{0.8366}                        & \textbf{0.8342}                        & \textbf{0.8364}                        & \textbf{0.8889}                        \\
				LWF                               & {\color[HTML]{3166FF} \textbf{0.9628}} & 0.2654                                 & 0.2292                                 & 0.2544                                 & 0.2497                                 & 0.2973                                 & 0.3765                                 \\
				PNN                               & 0.8853                                 & 0.0796                                 & 0.2355                                 & 0.1997                                 & 0.2734                                 & 0.2673                                 & 0.3235                                 \\
				R\&R-Net (3 tasks)                & 0.9577                                 & {\color[HTML]{FE0000} \textbf{0.9356}} & 0.8967                                 & /                                      & /                                      & /                                      & /                                      \\
				R\&R-Net (6 tasks)                & 0.9452                                 & 0.9275                                 & 0.9014                                 & 0.7383                                 & 0.7281                                 & 0.7260                                 & 0.8277                                 \\
				NO-RL                             & {\color[HTML]{FE0000} \textbf{0.9663}} & 0.2073                                 & 0.2813                                 & 0.2844                                 & 0.3733                                 & 0.4253                                 & 0.4230                                 \\
				NO-RR                             & 0.9576                                 & {\color[HTML]{3166FF} \textbf{0.9349}} & {\color[HTML]{FE0000} \textbf{0.9081}} & {\color[HTML]{3166FF} \textbf{0.7907}} & {\color[HTML]{3166FF} \textbf{0.7823}} & {\color[HTML]{3166FF} \textbf{0.7747}} & {\color[HTML]{3166FF} \textbf{0.8580}} \\
				Proposed (SILF)                   & 0.9610                                 & 0.9339                                 & {\color[HTML]{3166FF} \textbf{0.9065}} & {\color[HTML]{FE0000} \textbf{0.8094}} & {\color[HTML]{FE0000} \textbf{0.7924}} & {\color[HTML]{FE0000} \textbf{0.7914}} & {\color[HTML]{FE0000} \textbf{0.8658}} \\ \toprule[1pt]
			\end{tabular}
		}
	\end{center}
\end{table*}

\subsubsection{Implementation Details}
Our proposed SILF is implemented with the PyTorch library. All experiments are executed on Ubuntu 18.04 with i9-9900k CPU and two NVIDIA GTX TITAN XP GPUs. We utilize ResNeXt101 \cite{xie2017aggregated} as the backbone network of our SILF, and we replace the last two fully connected layers of ResNeXt101 with one fully connected layer of size [2048, 1]. The parameters of the backbone network are initialized by the weights pre-trained on ImageNet \cite{deng2009imagenet}. We add a sigmoid activation function after the fully connected layer to make the score output by the network in the range [0, 1]. We set the number of preset and additional tasks to 3 and the first and second pruning ratios to [0.7, 0.5, 0] and [0.4, 0.4, 0.4], respectively. We adopt the $L_{1}$ loss as the loss function for each task and apply stochastic gradient descent (SGD) as the optimizer. We set the initial learning rate to $1 \times 10^{-3}$ with a decay factor of 0.5 for every 10 epochs. 
Following previous works \cite{zhang2018blind, su2020blindly}, we split 80\% of each dataset into training sets and the remaining 20\% into test sets without using validation sets. We stop training at a preset 40 epoch.
During training, we crop each image into five $256 * 256$ patches (center and four corners) and take the average score of these patches as the final score for this image. Following the similar setting of \cite{liu2022liqa}, we repeat the random split three times on each database and report the average SRCC result across all trials.

\subsection{Comparison Methods}
We compared six methods, including two benchmarks and four incremental learning strategies. The same backbone network is applied to each method for a fair comparison. The detailed settings of these methods are presented in the following:

\begin{itemize}
	\item{Separate Learning (SL)}: Each task is trained with an independent model. The results are usually considered an upper bound on the performance of BIQA.
	\item{NO Remember Learning (NO-RL)}: Retrain new tasks directly on the old model, where all tasks share all parameters of the same model. This approach usually leads to catastrophic forgetting.
	\item{NO Relevance Reuse (NO-RR)}: We set the value of $\lambda$ in (\ref{equation_3}) to $0$. In this setting, the relevance between tasks is not considered during training. The rest are the same as SILF.
	\item{Learning without Forgetting (LwF)} \cite{li2017learning}: LWF incorporates knowledge distillation and fine-tuning to learn new tasks while retaining outputs of old tasks. For each new task, we add $256$ neurons to the penultimate fully connected layer of the network.
	\item{Progressive Neural Networks (PNN)} \cite{rusu2016progressive}: PNN keeps a library of models during training and adds lateral connections between models to extract features useful for new tasks. We add a branch for each new task and use lateral connections between branches.
	\item{Remember and Reuse Network (R\&R-Net)} \cite{ma2021remember}: We assign parameters equally to each task. It is worth noting that this method requires us to set the number of tasks in advance. In practice, R\&R-Net is unable to learn additional tasks.
\end{itemize}

\begin{table*}[]
	\begin{center}
		\caption{Cross-Task BIQA performance comparison (SRCC) on task sequence I}
		\label{table_4}
		\setlength{\tabcolsep}{4mm}{
			\begin{tabular}{c|c|l|cccccc}
				\toprule[1pt]
				\textbf{Sequence}           & \textbf{Dataset}                   & \textbf{Method} & \textbf{LIVE-CH} & \textbf{CSIQ}   & \textbf{LIEQ}   & \textbf{KONIQ}  & \textbf{LIVE-MD} & \textbf{SHRQ-A} \\ \toprule[1pt]
				-                           & \textbf{All}                       & SL              & 0.8529            & 0.9377          & 0.9064          & 0.8991          & 0.9564            & 0.9620           \\ \hline
				\multirow{6}{*}{\textbf{$1^{st}$}} & \multirow{6}{*}{\textbf{LIVE-CH}} & LWF             & 0.8260            & -               & -               & -               & -                 & -                \\
				&                                    & PNN             & 0.6809            & -               & -               & -               & -                 & -                \\
				&                                    & R\&R-Net        & 0.7910            & -               & -               & -               & -                 & -                \\
				&                                    & NO-RL           & 0.8039            & -               & -               & -               & -                 & -                \\
				&                                    & NO-RR           & \textbf{0.8494}   & -               & -               & -               & -                 & -                \\
				&                                    & Proposed (SILF) & 0.8472            & -               & -               & -               & -                 & -                \\ \hline
				\multirow{6}{*}{\textbf{$2^{nd}$}} & \multirow{6}{*}{\textbf{CSIQ}}     & LWF             & 0.6621            & 0.0971          & -               & -               & -                 & -                \\
				&                                    & PNN             & -0.1681           & 0.7998          & -               & -               & -                 & -                \\
				&                                    & R\&R-Net        & 0.7910            & 0.8807          & -               & -               & -                 & -                \\
				&                                    & NO-RL           & 0.0668            & 0.8658          & -               & -               & -                 & -                \\
				&                                    & NO-RR           & \textbf{0.8494}   & 0.9077          & -               & -               & -                 & -                \\
				&                                    & Proposed (SILF) & 0.8472            & \textbf{0.9086} & -               & -               & -                 & -                \\ \hline
				\multirow{6}{*}{\textbf{$3^{rd}$}} & \multirow{6}{*}{\textbf{LIEQ}}     & LWF             & 0.6599            & -0.2293         & 0.6389          & -               & -                 & -                \\
				&                                    & PNN             & 0.0446            & 0.1965          & 0.5510          & -               & -                 & -                \\
				&                                    & R\&R-Net        & 0.7910            & 0.8807          & \textbf{0.8963} & -               & -                 & -                \\
				&                                    & NO-RL           & 0.1920            & 0.2592          & 0.8852          & -               & -                 & -                \\
				&                                    & NO-RR           & \textbf{0.8494}   & 0.9077          & 0.8512          & -               & -                 & -                \\
				&                                    & Proposed (SILF) & 0.8472            & \textbf{0.9086} & 0.8481          & -               & -                 & -                \\ \hline
				\multirow{6}{*}{\textbf{$4^{th}$}} & \multirow{6}{*}{\textbf{KONIQ}}    & LWF             & 0.7706            & -0.5298         & 0.6326          & 0.7365          & -                 & -                \\
				&                                    & PNN             & 0.7428            & -0.6403         & 0.5367          & 0.8665          & -                 & -                \\
				&                                    & R\&R-Net        & 0.7910            & 0.8807          & \textbf{0.8963} & 0.8694          & -                 & -                \\
				&                                    & NO-RL           & 0.7973            & -0.7429         & 0.6487          & \textbf{0.8973} & -                 & -                \\
				&                                    & NO-RR           & 0.7929            & 0.8880          & 0.7830          & 0.8916          & -                 & -                \\
				&                                    & Proposed (SILF) & \textbf{0.8045}   & \textbf{0.8984} & 0.8343          & 0.8902          & -                 & -                \\ \hline
				\multirow{6}{*}{\textbf{$5^{th}$}} & \multirow{6}{*}{\textbf{LIVE-MD}} & LWF             & 0.6347            & 0.1613          & 0.3391          & 0.5302          & 0.1060            & -                \\
				&                                    & PNN             & 0.1820            & 0.4289          & 0.1869          & 0.1097          & 0.4556            & -                \\
				&                                    & R\&R-Net        & 0.7910            & 0.8807          & \textbf{0.8963} & 0.8694          & 0.6657            & -                \\
				&                                    & NO-RL           & 0.2769            & 0.5638          & -0.1346         & 0.3627          & 0.9354            & -                \\
				&                                    & NO-RR           & 0.7929            & 0.8880          & 0.7830          & \textbf{0.8916} & 0.9402            & -                \\
				&                                    & Proposed (SILF) & \textbf{0.8045}   & \textbf{0.8984} & 0.8343          & 0.8902          & \textbf{0.9553}   & -                \\ \hline
				\multirow{6}{*}{\textbf{$6^{th}$}} & \multirow{6}{*}{\textbf{SHRQ-A}}  & LWF             & 0.6280            & -0.2274         & 0.4695          & 0.5440          & -0.3616           & 0.7411           \\
				&                                    & PNN             & 0.3803            & -0.2211         & 0.5078          & 0.4398          & -0.0952           & 0.8145           \\
				&                                    & R\&R-Net        & 0.7910            & 0.8807          & \textbf{0.8963} & 0.8694          & 0.6657            & 0.8478           \\
				&                                    & NO-RL           & 0.5931            & -0.7045         & 0.5703          & 0.5866          & -0.2739           & \textbf{0.8935}  \\
				&                                    & NO-RR           & 0.7929            & 0.8880          & 0.7830          & \textbf{0.8916} & 0.9402            & 0.7565           \\
				&                                    & Proposed (SILF) & \textbf{0.8045}   & \textbf{0.8984} & 0.8343          & 0.8902          & \textbf{0.9553}   & 0.8045           \\ 
				\toprule[1pt]
			\end{tabular}
		}
	\end{center}
\end{table*}

\subsection{Main Results}
To evaluate the cross-task BIQA performance of different methods, we divide the twelve tasks into two task sequences and conduct experiments on them. Tables \ref{table_2} and \ref{table_3} show the average accuracy results for Sequence I and Sequence II, respectively. The results in Tables \ref{table_2} and \ref{table_3} first row are for separate learning (SL), which is usually considered the upper bound of BIQA performance and is marked in bold. In addition, we employ red and blue fonts to indicate the first and second ranking of performance among the remaining methods, respectively. 

There are several valuable findings from the analysis of Tables \ref{table_2} and \ref{table_3}. First, NO-RL performance on learned tasks decreased significantly with the sequential learning process due to the inability to remember learned tasks (i.e., experienced catastrophic forgetting during incremental learning). Second, LWF and PNN are usually used for classification scenarios that are relatively similar between tasks. The cross-task BIQA results of LWF and PNN are unsatisfactory due to the significant differences among tasks in cross-task BIQA. Third, R\&RNet shows a remarkable ability to avoid forgetting. Still, its average accuracy on multiple tasks is lower than NO-RR and SILF because it employs only fixed first pruning. Last, our proposed SILF can effectively exploit the inter-task relevance information to improve the performance of cross-task BIQA compared to NO-RR. In Tables \ref{table_2} and \ref{table_3}, the mean values of SILF average accuracy are improved by $1.05\%$ and $0.91\%$ compared to NO-RR.

Table \ref{table_4} shows the cross-task BIQA results for task sequence I. The second and third columns in Table \ref{table_4} show the sequential training datasets and the comparison methods, respectively. The model for each method was tested on all learned tasks as each new task was completed. The test results are SRCC values, as shown in Table \ref{table_4}. The second row of Table \ref{table_4} shows the results of separate learning (SL), which can be considered an upper bound on the performance of each task. Rows 3-9 in Table \ref{table_4} show the results of each method learned sequentially and tested on learned tasks.

\begin{table*}[]
	\begin{center}
		\caption{Different training orders. A, S, and E are acronyms for Authentic, Synthetic, and Enhancement respectively}
		\label{table_5}
		\setlength{\tabcolsep}{4.6mm}{
			\begin{tabular}{l|cccccc}
				\toprule[1pt]
				\textbf{Order} & \textbf{$1^{st}$} & \textbf{$2^{nd}$} & \textbf{$3^{rd}$} & \textbf{$4^{th}$} & \textbf{$5^{th}$} & \textbf{$6^{th}$}  \\ \toprule[1pt]
				I              & A (LIVE-CH) & S (CSIQ)    & E (LIEQ)    & A (KONIQ)   & S (LIVE-MD) & E (SHRQ-A)  \\
				II             & A (LIVE-CH) & E (LIEQ)    & S (CSIQ)    & A (KONIQ)   & E (SHRQ-A)  & S (LIVE-MD) \\
				III            & S (CSIQ)    & A (LIVE-CH) & E (LIEQ)    & S (LIVE-MD) & A (KONIQ)   & E (SHRQ-A)  \\
				IV             & S (CSIQ)    & E (LIEQ)    & A (LIVE-CH) & S (LIVE-MD) & E (SHRQ-A)  & A (KONIQ)   \\
				V              & E (LIEQ)    & A (LIVE-CH) & S (CSIQ)    & E (SHRQ-A)  & A (KONIQ)   & S (LIVE-MD) \\
				VI             & E (LIEQ)    & S (CSIQ)    & A (LIVE-CH) & E (SHRQ-A)  & S (LIVE-MD) & A (KONIQ)   \\ \toprule[1pt]
			\end{tabular}
		}
	\end{center}
\end{table*}

\subsection{Ablation Experiments}
In this subsection, we conduct a series of ablation experiments to evaluate the factors that affect the performance of our method.

\subsubsection{The Impact of Training Order}
To evaluate the robustness of our method to the training order, we adjusted the order of the tasks in Sequence I. Sequence I contains two tasks for each of the three task types. Please refer to Table \ref{table_1} for the specific distortion type of each task. By permuting three different distortion types, we obtained six different training orders for Sequence I, as shown in Table \ref{table_5}.

\begin{table}[h]
	\begin{center}
		\caption{Ablation experimental results about the training order}
		\label{table_6}
		\setlength{\tabcolsep}{4mm}{
			\begin{tabular}{l|ccc}
				\toprule[1pt]
				\textbf{\begin{tabular}[c]{@{}c@{}}Training\\ Order\end{tabular}} & \textbf{\begin{tabular}[c]{@{}c@{}}Average\\ Accuracy\end{tabular}} & \textbf{\begin{tabular}[c]{@{}c@{}}Average\\ Plasticity\end{tabular}} & \textbf{\begin{tabular}[c]{@{}c@{}}Average\\ Forgetting\end{tabular}} \\ \toprule[1pt]
				I              & 0.8652                                 & {\color[HTML]{FE0000} \textbf{0.8757}} & 0.0104                                 \\
				II             & {\color[HTML]{FE0000} \textbf{0.8698}} & 0.8686                                 & 0.0077                                 \\
				III            & 0.8651                                 & {\color[HTML]{3166FF} \textbf{0.8578}} & {\color[HTML]{3166FF} \textbf{0.0186}} \\
				IV             & 0.8685                                 & 0.8594                                 & {\color[HTML]{FE0000} \textbf{0.0072}} \\
				V              & {\color[HTML]{3166FF} \textbf{0.8442}} & 0.8667                                 & 0.0087                                 \\
				VI             & 0.8606                                 & 0.8754                                 & 0.0088                                 \\ \toprule[1pt]
			\end{tabular}
		}
	\end{center}
\end{table}

Table \ref{table_6} reports the results of the ablation experiments with different training orders. Each metric's best and worst values are marked in red and blue, respectively. We apply the range of fluctuations and standard deviations to measure the robustness of our proposed model to the training order. The difference between the three metrics' maximum and minimum values are $0.0256$, $0.0179$, and $0.0114$, and their standard deviations are $0.0094$, $0.0076$, and $0.0042$, respectively. We can infer from the above analysis that our proposed model is insensitive to the training order and has high robustness.

\subsubsection{The Impact of Second Pruning Ratio}
Table \ref{table_7} shows the performance of our proposed model under different second pruning ratio settings. The top three values of each metric are marked in red, green, and blue. 

We can summarize the following points from Table \ref{table_7}. First, the average accuracy takes higher values at second pruning ratios of $0.2$, $0.3$, and $0.4$, decreasing significantly when the second pruning ratio is $0.6$. When the second pruning ratio is $0.7$, the model cannot infer due to insufficient remaining parameters. Second, average plasticity and average forgetting were positively correlated with the second pruning ratio. Considering the two points mentioned above, we take $0.4$ as the second pruning ratio of our proposed model.

\begin{table}[t]
	\begin{center}
		\caption{Ablation experimental results about the second pruning ratio}
		\label{table_7}
		\setlength{\tabcolsep}{3mm}{
			\begin{tabular}{l|ccc}
				\toprule[1pt]
				\multicolumn{1}{c|}{\textbf{\begin{tabular}[c]{@{}c@{}}Second Pruning\\ Ratio\end{tabular}}} & \textbf{\begin{tabular}[c]{@{}c@{}}Average \\ Accuracy\end{tabular}} & \textbf{\begin{tabular}[c]{@{}c@{}}Average \\ Plasticity\end{tabular}} & \textbf{\begin{tabular}[c]{@{}c@{}}Average \\ Forgetting\end{tabular}} \\ \toprule[1pt]
				0.1                                                                                        & 0.8562                                                               & 0.7935                                                                 & {\color[HTML]{FE0000} \textbf{0.0029}}                                 \\
				0.2                                                                                        & {\color[HTML]{FE0000} \textbf{0.8662}}                               & 0.8366                                                                 & {\color[HTML]{32CB00} \textbf{0.0035}}                                 \\
				0.3                                                                                        & {\color[HTML]{3166FF} \textbf{0.8632}}                               & 0.8498                                                                 & {\color[HTML]{3166FF} \textbf{0.0051}}                                 \\
				0.4 (SILF)                                                                                 & {\color[HTML]{32CB00} \textbf{0.8652}}                               & {\color[HTML]{32CB00} \textbf{0.8757}}                                 & 0.0104                                                                 \\
				0.5                                                                                        & 0.8514                                                               & {\color[HTML]{3166FF} \textbf{0.8748}}                                 & 0.0301                                                                 \\
				0.6                                                                                        & 0.6633                                                               & {\color[HTML]{FE0000} \textbf{0.8855}}                                 & 0.3326                                                                 \\
				0.7                                                                                        & -                                                                    & -                                                                      & -                                                                      \\ \toprule[1pt]
			\end{tabular}
		}
	\end{center}
\end{table}

\begin{table}[t]
	\begin{center}
		\caption{Ablation experimental results about the relevant constant}
		\label{table_8}
		\setlength{\tabcolsep}{4mm}{
			\begin{tabular}{l|ccc}
				\toprule[1pt]
				\multicolumn{1}{c|}{\textbf{\begin{tabular}[c]{@{}c@{}}Relevant\\ Constant\end{tabular}}} & \textbf{\begin{tabular}[c]{@{}c@{}}Average\\ Accuracy\end{tabular}} & \textbf{\begin{tabular}[c]{@{}c@{}}Average\\ Plasticity\end{tabular}} & \textbf{\begin{tabular}[c]{@{}c@{}}Average\\ Forgetting\end{tabular}} \\ \toprule[1pt]
				0 (NO-RR)                                                                                  & 0.8562                                                              & 0.8661                                                                & 0.0226                                                                \\
				0.1                                                                                        & {\color[HTML]{32CB00} \textbf{0.8636}}                              & 0.8614                                                                & {\color[HTML]{3166FF} \textbf{0.0117}}                                \\
				0.3                                                                                        & 0.8547                                                              & 0.8474                                                                & {\color[HTML]{FE0000} \textbf{0.0078}}                                \\
				0.5 (SILF)                                                                                 & {\color[HTML]{FE0000} \textbf{0.8652}}                              & {\color[HTML]{FE0000} \textbf{0.8757}}                                & {\color[HTML]{32CB00} \textbf{0.0104}}                                \\
				0.7                                                                                        & {\color[HTML]{3166FF} \textbf{0.8564}}                              & {\color[HTML]{32CB00} \textbf{0.8752}}                                & 0.0125                                                                \\
				0.9                                                                                        & 0.8183                                                              & {\color[HTML]{3166FF} \textbf{0.8737}}                                & 0.0714                                                                \\ \toprule[1pt]
			\end{tabular}
		}
	\end{center}
\end{table}

\subsubsection{The Impact of Relevant Constant}
Here, we evaluate the effect of $\lambda$ in (\ref{equation_3}). Table \ref{table_8} demonstrates the performance of our proposed model under different relevant constant settings. The top three values of each metric are marked in red, green, and blue. 

Table \ref{table_8} shows that the average accuracy and average plasticity metrics are both highest at a relevant constant of $0.5$, and the average forgetting metric is the second highest. In addition, our proposed SILF improves by approximately $1.05\%$, $1.11\%$, and $1.22\%$ for the three metrics, respectively, compared to the relevant constant taking a value of 0 (i.e., NO-RR).

\section{Conclusion}
The R\&R-Net proposed in our previous work addresses catastrophic forgetting in cross-task BIQA. However, R\&R-Net cannot learn additional tasks due to a rigid parameter assignment strategy. This paper presents a scalable incremental learning framework (SILF) to extend R\&R-Net by utilizing a scalable incremental learning strategy. Specifically, by gradually and selectively pruning unimportant neurons from previously settled parameter subsets, we develop a scalable memory unit, which enables us to forget part of previous experiences and free the limited memory capacity for adapting to emerging new tasks. Extensive experiments on eleven IQA datasets demonstrate that our proposed method significantly outperforms the other state-of-the-art methods in cross-task BIQA.

\bibliographystyle{IEEEtran}
\bibliography{reference}

\end{document}